\documentclass[lettersize,journal]{IEEEtran}
\usepackage{amsmath,amsfonts}
\usepackage{algorithmic}
\usepackage{algorithm}
\usepackage{array}
\usepackage[caption=false,fon  t=normalsize,labelfont=sf,textfont=sf]{subfig}
\usepackage{textcomp}
\usepackage{stfloats}
\usepackage{url}
\usepackage{verbatim}
\usepackage{graphicx}
\usepackage{booktabs}
\usepackage{multirow} 
\usepackage{cite}
\usepackage[pagebackref=true,breaklinks=true,colorlinks,bookmarks=false,citecolor=blue,linkcolor=blue]{hyperref}
\usepackage{bm}

\begin{document}

\title{LaneTCA: Enhancing Video Lane Detection with Temporal Context Aggregation}

\author{
 Keyi Zhou,
 Li Li,
 Wengang Zhou,~\IEEEmembership{Senior Member,~IEEE}, 
 Yonghui Wang,
 Hao Feng,
 \par
 and~Houqiang Li,~\IEEEmembership{Fellow,~IEEE}

\IEEEcompsocitemizethanks{
\IEEEcompsocthanksitem Keyi Zhou, Li Li, Wengang Zhou, Yonghui Wang, Hao Feng, and Houqiang Li are with the National Engineering Laboratory for Brain-Inspired Intelligence Technology and Application, Department of Electronic Engineering and Information Science, University of Science and Technology of China, Hefei, 230027, China.
E-mail: \{kyzhou2000, wyh1998, haof\}@mail.ustc.edu.cn; \{lil1, zhwg, lihq\}@ustc.edu.cn
}}

\maketitle

\begin{abstract}
In video lane detection, there are rich temporal contexts among successive frames, 
which is under-explored in existing lane detectors.
In this work, we propose LaneTCA to bridge the individual video frames and explore how to effectively aggregate the temporal context. Technically, we develop an accumulative attention module and an adjacent attention module to abstract the long-term and short-term temporal context, respectively. 
The accumulative attention module continuously accumulates visual information during the journey of a vehicle, while the adjacent attention module propagates this lane information from the previous frame to the current frame.
The two modules are meticulously designed based on the transformer architecture. 
Finally, these long-short context features are fused with the current frame features to predict the lane lines in the current frame.
Extensive quantitative and qualitative experiments are conducted on two prevalent benchmark datasets. 
The results demonstrate the effectiveness of our method, achieving several new state-of-the-art records. The codes and models are available at \href{https://github.com/Alex-1337/LaneTCA}{{\tt https://github.com/Alex-1337/LaneTCA}}
\end{abstract}


\section{Introduction}

\begin{figure}[t]
    \centering
    \includegraphics[width=1\columnwidth]{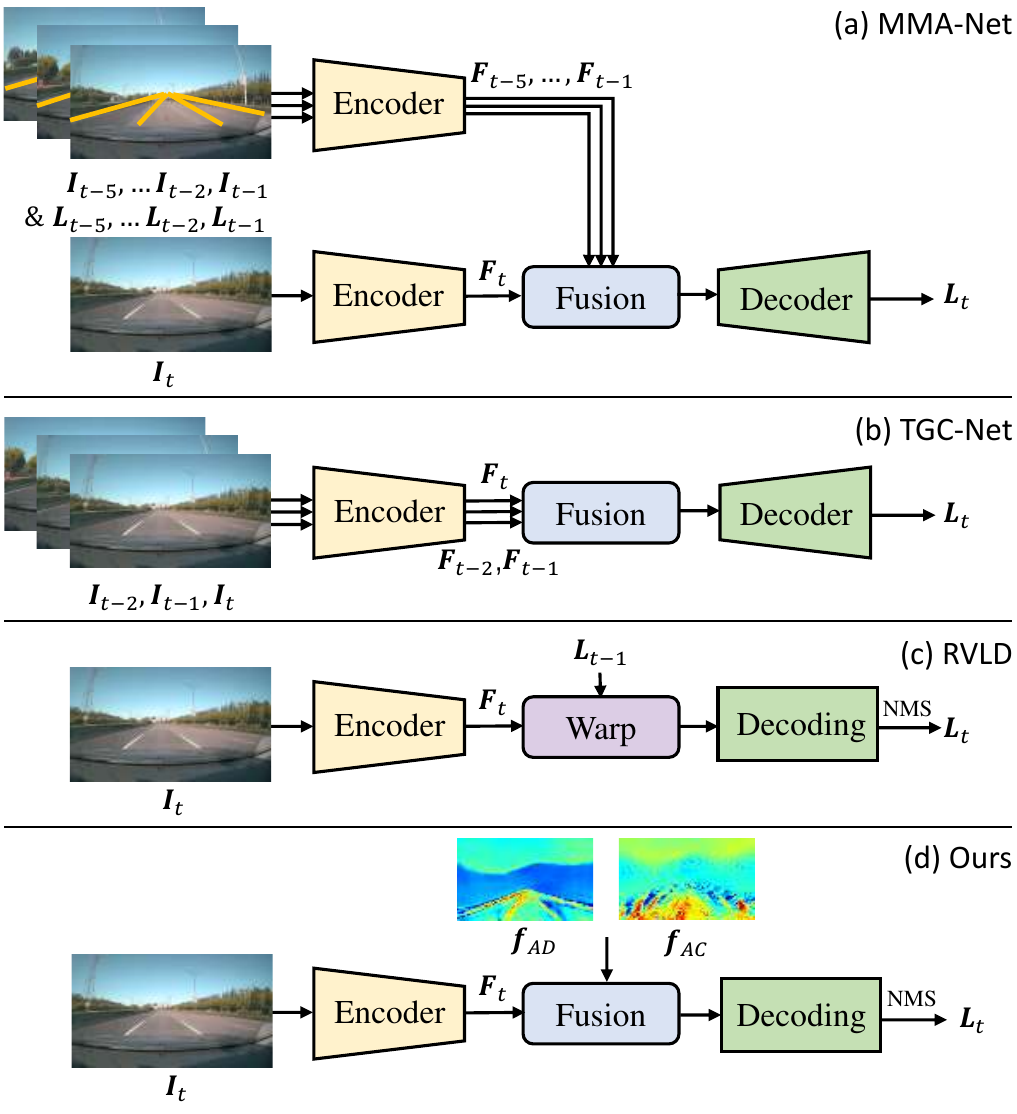}
    \caption{Comparison of existing video lane detection methods: (a) MMA-Net~\cite{zhang2021vil}, requiring multiple frames of images and detection results, (b) TGC-Net, requiring multiple frames of images, (c) RVLD~\cite{jin2023recursive}, requiring the current frame of image and the previous frame of detection results, and (d) our proposed LaneTCA, requiring the current frame $\bm{I}_t$, adjacent feature $\bm{f}_{AD}$, and accumulative feature $\bm{f}_{AC}$.  LaneTCA requires only the visual information from the current frame while also considering information from different time spans. Concretely, $\bm{f}_{AD}$ contains information from $\bm{I}_{t-1}$, while $\bm{f}_{AC}$ contains information from frames $\bm{I}_0$ to $\bm{I}_{t-1}$. 
    }
    \label{fig:intro}
\end{figure}

Lane detection is crucial for autonomous driving~\cite{levinson2011towards, wu2023dense, yuan2021temporal, tao2023pseudo}, as it allows the identification of lane boundaries and facilitates the determination of secure driving routes~\cite{bar2014recent, satzoda2015enhancing, wu2023dense}. 
The RGB images captured by vehicle-mounted cameras provide multiple forward lane lines and other useful driving information~\cite{zhang2022online,natarajan2015multi}.
In recent years, with the emergence of various methods~\cite{grant2017crowd,neven2018towards,li2019line,lu2022bidirectional,mithun2020rgb2lidar,chu2021neighbor}, lane detection has seen significant advancements.
However, this task presents various challenges, including discontinuous lane markings, occlusions caused by vehicles or pedestrians, and environmental issues such as glare or nighttime conditions. 

Early lane detection methods are image-based, commonly used edge decomposition and Hough transform-based techniques~\cite{illingworth1988survey,mukhopadhyay2015survey} to locate lane markings in the parameter space~\cite{liu2010combining,zhou2010novel}. However, these methods have limited adaptability to complex scenes and heavily rely on parameter settings.
In recent years, deep learning-based methods~\cite{pan2018spatial, li2019line, tabelini2021polylanenet} have emerged as a promising alternative solutions, achieving remarkable results. Some approaches are focused on segmentation~\cite{pan2018spatial,zheng2021resa,xu2020curvelane}, learning feature vectors for each position to delineate each lane marking. 
Unlike segmentation-based methods, anchor-based lane detection methods~\cite{li2019line,tabelini2021keep,zheng2022clrnet,qin2022ultra,liu2021condlanenet} detect lane lines as a whole, alleviating detection errors caused by occlusion. Furthermore, curve-based methods~\cite{tabelini2021polylanenet,liu2021end,feng2022rethinking} have contributed to the lane detection task by only requiring the prediction of the curve parameters to obtain the lane lines. 
Although the image-based detection techniques work well, they ignore the continuity between frames in video sequences which provides additional valuable information for lane detection in autonomous driving scenarios, as discussed next. 

Video lane detection leverages information from historical frame sequences, expanding upon single-frame lane detection methods to accommodate video-based tasks. Various approaches design networks to better integrate historical information, with the aim of achieving more accurate detection results for the current frame.
Figure~\ref{fig:intro} illustrates the comparison of existing video lane detection pipelines.  
As shown in Figure~\ref{fig:intro}(a), methods such as MMA-Net~\cite{zhang2021vil} provide rich reference information, requiring multiple historical frames and their corresponding lane line masks as input for each detection frame. However, such methods require complex visual information for each frame. This leads to the issue that if the predictions for historical frames are incorrect, they will directly impact the prediction for the current frame.
Some methods, represented by TGC-Net~\cite{wang2022video}, require inputting a fixed number of frames of multi-frame images for information fusion, as shown in Figure~\ref{fig:intro}(b). They do not require detection results from previous frames as input. However, the fixed frame number setting limits reference in terms of time span.

In contrast to above methods that require multiple frames of images, the latest approach, RVLD~\cite{jin2023recursive} in Figure~\ref{fig:intro}(c), utilizes warping method, requiring only the prediction of previous frame as historical information to reach the state-of-the-art results. However, this method may be inadequate for long-term information utilization.
In conclusion, it can be seen that the core of video tasks lies in how to better integrate multi-frame information or transform information from the previous frame into the current frame's results. Nevertheless, many of these methods face challenges such as complex input or insufficient use of historical information.

To address the aforementioned issues, we propose a novel video lane detection method with a Temporal Context Aggregation network, namely LaneTCA. As shown in Figure~\ref{fig:intro}(d), it can utilize the previous frame as direct historical information while referencing all prior frames to provide a more comprehensive implicit historical context. Concretely, our approach incorporates a temporal context aggregation network for updating historical information, which includes adjacent attention and accumulative attention mechanisms. Adjacent attention leverages highly relevant information from the preceding frame, while accumulative attention continuously learns road information from the beginning of the video, accumulating all road features from the first frame onward. Consequently, the accumulative attention structure naturally provides long time span reference information even when only a single query is input. 
After obtaining the aforementioned short-term and long-term features, LaneTCA integrates these with the initial features directly obtained from the current frame. The optimized features are then decoded to produce the final lane line markings. 

To verify our LaneTCA, we perform extensive experiments on prevalent challenging
benchmark datasets. The quantitative and qualitative results exhibit the effectiveness of our method. The contributions of our method are summarized as follows:
\begin{itemize}
\item 
We propose LaneTCA, a novel framework 
for video lane detection with effective temporal context aggregation.

\item 
We introduce accumulative and adjacent attention modules for long-range and short-range temporal information. This approach transcends fixed frame limitations, advancing beyond previous multi-frame methods.

\item We perform extensive experiments on prevalent benchmark datasets. The experimental results demonstrate the effecitveness and its superior performance.

\end{itemize}

\section{Related Work}

Lane detection tasks can be categorized into two main areas, which are image-based lane detection and video-based lane detection. In the following section, we first present a concise overview of the historical development of these two categories, followed by a brief overview of the historical development of the attention mechanisms.

\subsection{Image Lane Detection}
Based on the representation of lane lines, image lane detection tasks can be classified into three categories: segmentation-based, anchor-based, and curve-based methods. In the following, we provide a brief overview of the historical development of these categories.

Segmentation-based methods perform pixel-wise segmentation of lanes. SCNN~\cite{pan2018spatial} uses slice-by-slice convolution to horizontally and vertically pass feature map information and obtain global feature aggregation, but it is slow for real-time detection. To address this issue, RESA~\cite{zheng2021resa} improves the speed and performance significantly by recurrent feature-shift. CurveLane-NAS~\cite{xu2020curvelane} significantly improves the performance by considering variations in the shape of lane lines and the importance of information from both long and short distances of lane lines. NAS is applied to optimize the entire system, but it relies on substantial computational resources, making it quite time-consuming.

The anchor-based methods can be classified into those based on line anchor and those based on row anchor. The first line anchor-based method, Line-CNN~\cite{li2019line}, utilizes line proposal to detect lane lines. LaneATT~\cite{tabelini2021keep}, another line anchor-based method, leverages the attention mechanism to achieve high performance and fast speed on multiple datasets. CLRNet~\cite{zheng2022clrnet}, a novel network architecture, employs features from both low-level and high-level to achieve state-of-the-art results. In contrast, row anchor-based methods determine the location of points by predicting the cell location in each row of the image. UFLD~\cite{qin2022ultra} introduces a hybrid anchor system using both row and column anchors, while CondLaneNet~\cite{liu2021condlanenet} adopts a conditional shape head to predict the row-wise location of lanes. A potential disadvantage of the point sampling-based approach is the reliance on NMS~\cite{bodla2017soft} post-processing.

Lane detection methods based on curve detection present lanes as curves with parameters, enabling high-speed detection by only regressing these parameters. For example, PolyLaneNet~\cite{tabelini2021polylanenet} uses a third-order polynomial to represent lanes, achieving high efficiency with a simple approach. LSTR~\cite{liu2021end} leverages transformer~\cite{vaswani2017attention} to incorporate global information for better accuracy, and also considers road structures and camera pose as parameters. BézierLaneNet~\cite{feng2022rethinking} replaces the polynomial with the classic cubic Bézier curve~\cite{choi2008path,liu2020abcnet}, which shows high stability and flexibility in complex driving scenes, benefiting high-speed performance.

\begin{figure*}[t]
    \centering
    \includegraphics[width=2\columnwidth]{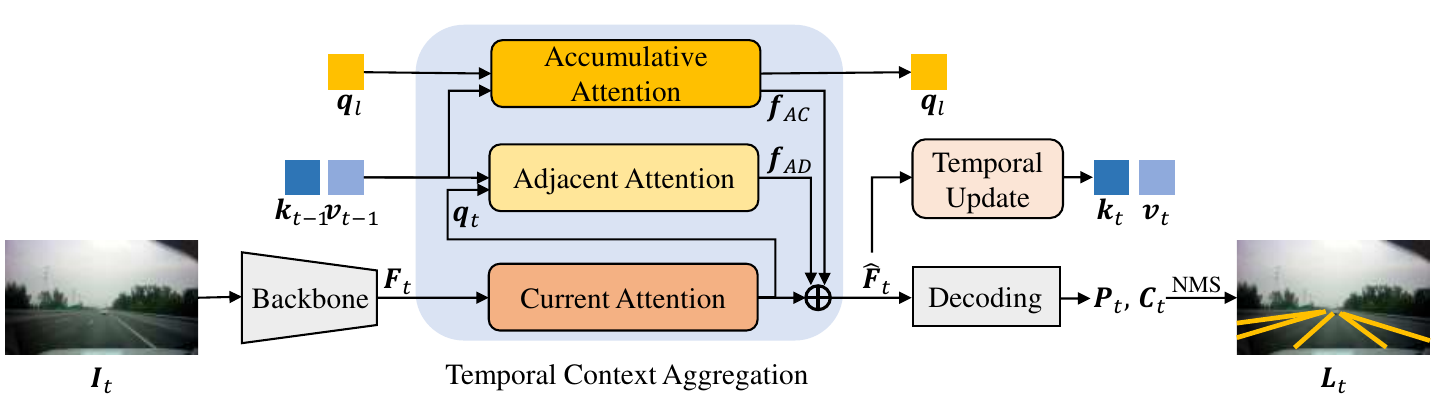}
    \caption{An overview of the proposed LaneTCA. With the given image $\bm{I}_t$, we first apply an encoder to extract features $\bm{F}_t$. 
    The features of the current frame are fed into the current attention module in the temporal context aggregation network.
    The key $\bm{k}_{t-1}$ and the value $\bm{v}_{t-1}$ from the previous frame with the output of the current attention module are fed into the adjacent attention module.
    A learnable query $\bm{q}_l$ along with $\bm{k}_{t-1}$ and $\bm{v}_{t-1}$ is input into the accumulative attention module.
    The outputs from these three channels are combined to obtain the optimized current frame information.
    The optimized output features are decoded through a series of steps to obtain the probability map $\bm{P}_{t}$ and parameter map $\bm{C}_{t}$, and then processed with NMS to produce the final lane lines $\bm{L}_t$. 
    The optimized features provide $\bm{k}_{t}$ and $\bm{v}_{t}$ for the current frame, 
    while the output of the accumulative channel continues to serve as $\bm{q}_l$.
    }
    \label{fig:framework}
\end{figure*}

\subsection{Video Lane Detection}

The development of video-based lane detection methods has progressed significantly over recent years. 
Early techniques, such as ConvLSTM~\cite{zou2019robust} and ConvGRUs~\cite{zhang2021lane}, employ recurrent neural networks to combine features from the current frame with those from previous frames, effectively leveraging temporal correlations. For example, MMA-Net~\cite{zhang2021vil}, which utilizes the first video lane dataset VIL-100, introduces an attention-based lane detection framework. This approach aggregates features from the current frame along with several preceding frames to improve lane identification accuracy. Similarly, LaneATT-T~\cite{tabelini2022lane} extracts multi-frame lane features using a line anchor detector, while TGC-Net~\cite{wang2022video} advances spatial and temporal information extraction between adjacent frames by extending the feature aggregation module from RESA. TGC-Net~\cite{wang2022video} also incorporates a specialized loss function to ensure geometric consistency of detected lanes.

Earlier methods typically aggregate features from multiple past frames.
In contrast, the state-of-the-art RVLD~\cite{jin2023recursive} method introduces a novel recursive approach that preserves information from a single previous frame, allowing for efficient propagation of temporal information. This approach enables more precise extraction of lane features in each frame by maintaining continuity and consistency over time.

Unlike previous methods, our proposed LaneTCA does not rely on a fixed number of historical frames and can effectively utilize prior information in subsequent frame predictions. Compared to methods that require long-term historical information, our proposed LaneTCA does not necessitate multiple frames of images as input each time, while still integrating information over both short and long time spans. Furthermore, in contrast to RVLD~\cite{jin2023recursive}, LaneTCA incorporates more extended historical information.

\subsection{Attention Mechanism}
In recent years, the attention mechanism~\cite{vaswani2017attention} has become pivotal in the realm of deep learning. Within natural language processing, the transformer~\cite{vaswani2017attention} architecture has showcased exceptional performance across various downstream tasks, largely due to its attention mechanism. Central to the transformer is the multi-head self-attention component, which processes a sequence of embeddings through linear projections and matrix multiplications to model their interdependencies. This mechanism enables the network to effectively capture the relationships between sequence elements, thereby enhancing its focus on critical embeddings. Motivated by the attention mechanism's success in NLP, the Vision Transformer~\cite{dosovitskiy2020image} introduced the transformer model to the visual domain, demonstrating its efficacy in this context. Since then, the attention mechanism has seen growing adoption in numerous computer vision tasks, such as object detection with DETR~\cite{carion2020end} and video instance segmentation with VisTR~\cite{wang2021end}. The attention mechanism within transformers has gained significant traction and has emerged as a competitive framework, rivaling convolutional networks across a variety of tasks~\cite{dosovitskiy2020image, carion2020end, wang2021end, liu2021swin}.

\section{Methodology}

In this section, we will sequentially introduce the architecture of the proposed LaneTCA, initialization setting, and training objective.

\subsection{Architecture of LaneTCA}
Figure~\ref{fig:framework} illustrates the architecture of our proposed LaneTCA. Given the current image and temporal information, our approach employs a backbone network to extract features from the current image, resulting in the initial features 
$\bm{F}_t\in \mathbb{R}^{H\times W\times C}$. The features $\bm{F}_t$ are then refined to $\hat{\bm{F}_t}\in \mathbb{R}^{H\times W\times C}$ using temporal information from the historical frames, enhancing the feature representation of the current frame. 
To obtain lane lines, 
the refined features $\hat{\bm{F}_t}$ are processed following the method in the eigenlanes\cite{jin2022eigenlanes}, where two decoding steps are performed to obtain a probability map and a parameter map, respectively. Finally, non-maximum suppression (NMS) is applied as a post-processing step to extract the lane lines.

\begin{figure}[t]
    \centering
    \includegraphics[width=1\columnwidth]{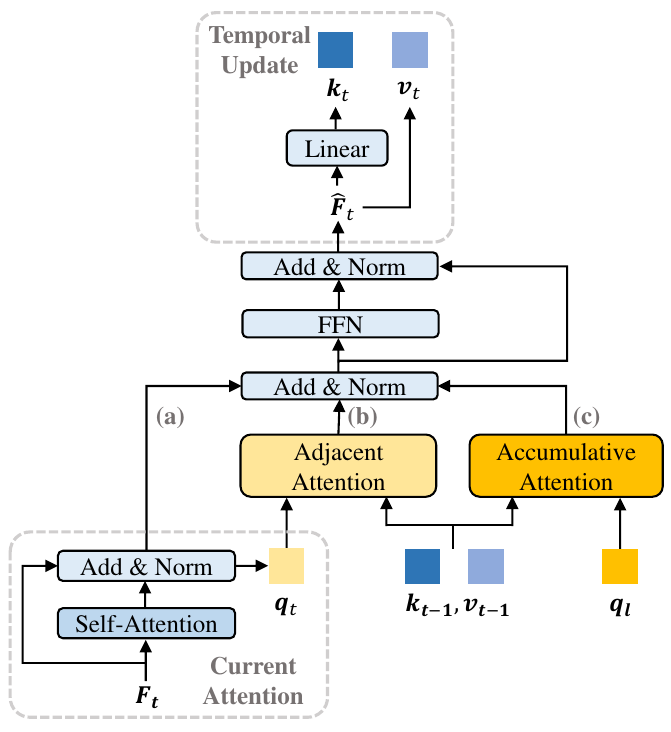}
    \caption{Illustration of the process of temporal context aggregation and temporal update. 
    The aggregation phase is divided into the following three branches: (a) current attention, (b) adjacent attention, and (c) accumulative attention.
    By inputting features $\bm{F}_t$ and the optimized $\bm{k}_{t-1}$ and $\bm{v}_{t-1}$ values from the branches produce individual outputs. These three attention results are then summed, obtaining a more accurately optimized feature
$\hat{\bm{F}_t}$. Subsequently, new 
$\bm{k}_t$ and 
$\bm{v}_t$ values are generated for the optimization of the next frame.
}
    \label{fig:refinement}
\end{figure}

\begin{figure}[t]
    \centering
    \includegraphics[width=1\columnwidth]{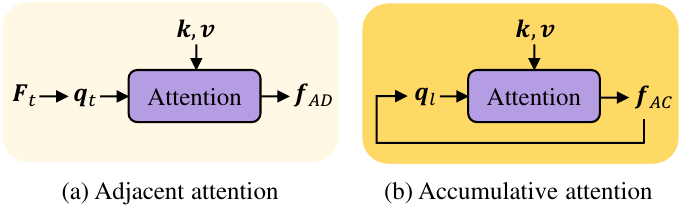}
    \caption{(a) Illustration of the adjacent attention module. The input $\bm{F}_t$ transform to $\bm{q}_t$ after self-attention operation. The attention operation is then conducted to obtain adjacent feature $\bm{f}_{AD}$. 
    (b) Illustration of the accumulative attention module. The $q_l$ is set as the learnable token. $\bm{f}_{AC}$ serves as $\bm{q}_l$ for the accumulative attention block in the subsequent frame.  
    }
    \label{fig:attention}
\end{figure}

The structure of temporal context aggregation network is shown in Figure~\ref{fig:refinement}. 
Overall, this network processes the input initial features $\bm{F}_t\in \mathbb{R}^{H\times W\times C}$ and temporal information.
Subsequently, self-attention, adjacent attention and accumulative attention are applied to obtain frame representations that reference information from the preceding frame and all historical frames, respectively. These representations are then aggregated with the initial features $\bm{f}_t\in \mathbb{R}^{H\times W\times C}$ of the current frame to obtain the final refined features $\hat{\bm{F}_t}\in \mathbb{R}^{H\times W\times C}$.

We divide the core of LaneTCA into the following modules: (a) current attention, (b) adjacent attention, (c) accumulative attention, and (d) temporal update. Each module will be described in detail in the subsequent subsections.

\smallskip
\textbf{Current attention.}
First, the feature $\bm{F}_t$  obtained from the encoder is used as the initial feature and input into the temporal context aggregation network. Then, after applying self-attention, we obtain $\bm{f}_t$. The feature $\bm{f}_t$ is subsequently transformed by a linear transformation $\phi$ to produce the query $\bm{q}_t\in \mathbb{R}^{HW\times C}$ for the current frame.

\smallskip
\textbf{Adjacent attention.} Lane markings typically do not undergo abrupt changes in morphology and structure, resulting in similar lane contours between adjacent frames. Therefore, information within a very short time span is beneficial in the task of video lane detection. Based on this consideration, we designed an adjacent attention module to leverage information from the previous frame as short-term memory, thereby enhancing the features extracted from the current frame.

Our adjacent attention module updates using the final optimized features obtained from the preceding frame. As illustrated in Figure~\ref{fig:attention}(a), the implementation involves combining $\bm{q}_t \in \mathbb{R}^{HW\times C}$ from the current frame with $\bm{k}_{t-1}\in \mathbb{R}^{HW\times C}$ and $\bm{v}_{t-1}\in \mathbb{R}^{HW\times C}$ from the previous frame for reference. The approach is expressed as follows:
\begin{equation}
\bm{f}_{AD} = Att(\bm{q}_t, \bm{k}_{t-1}, \bm{v}_{t-1}),
\end{equation}
where $\bm{f}_{AD}\in \mathbb{R}^{H\times W\times C}$ is the output of adjacent attention.

\smallskip
\textbf{Accumulative attention.} During vehicle operation, the lane lines over consecutive frames are often similar. However, due to the discontinuity of lane markings and occlusions caused by other vehicles or obstacles on the road, the adjacent images may sometimes differ significantly. Considering these factors, we believe that the information from multiple consecutive frames can complement each other. Referring to historical multi-frame information is likely to be more effective than only considering the previous frame. To incorporate longer-term information, we designed an accumulation attention.

In the accumulation attention module illustrated in Figure \ref{fig:attention} (b), we have implemented a learnable query $\bm{q}_l \in \mathbb{R}^{HW\times C}$ that undergoes continuous updates.
This learnable query $\bm{q}_l$ is set through zero initialization at the start of the video, specifically at the first frame.
Through these iterative updates, this query effectively encapsulates all historical information. The accumulative feature can be obtained by attention operation:
\begin{equation}
\bm{f}_{AC} = Att(\bm{q}_l, \bm{k}_{t-1}, \bm{v}_{t-1}),
\end{equation}
where $\bm{f}_{AC}\in \mathbb{R}^{H\times W\times C}$ is the output of accumulative attention block.

\smallskip
\textbf{Temporal update.}
The adjacent features $\bm{f}_{AD}$ are obtained through adjacent attention, while the global accumulative features $\bm{f}_{AC}$ are obtained through accumulative attention. These features are then combined with the features $\bm{f}_t$ that contain only the information of the current frame. This aggregation process results in optimized features $\hat{\bm{F}_t}$.

The optimized features are then processed through an additional mapping functions to obtain the updated $\bm{k}_t\in \mathbb{R}^{HW\times C}$ and pass $\hat{\bm{F}_t}$ as $\bm{v}_t\in \mathbb{R}^{HW\times C}$ for the current frame. The operation are shown as:
\begin{equation}
    \begin{aligned}
    &\bm{k}_t = \phi(\hat{\bm{F}_t}), \\
    &\bm{v}_t = \hat{\bm{F}_t}.\\
    \end{aligned}
\end{equation}
The obtained key-value pairs can be used for feature refinement in the next frame, continuously providing temporal information. That enhances the continuity of the predictions.

\smallskip
\subsection{Initialization Setting}
As a video task requiring the iterative use of temporal information, our method addresses the initialization problem: how to set the input when there is insufficient historical temporal information during the initial frames. The following section describes the specific approach.
For detection in the first frame, we follow an image-based lane detection approach to identify lane markings, using the features directly obtained from the encoder. For the second frame, the lane contour detected in the first frame is utilized as a mask to refine the initial features of the current frame.
  
After detecting the sampling points of lane line $\bm{L}_1=\left\{\bm{P}^1_1,...,\bm{P}^N_1 \right\} \in \mathbb{R}^{N\times2}$ in the first frame, we convert these points into a contour and then dilate them to a width of 1 pixel to generate the lane mask $\bm{M}$.

When starting the detection for the second frame, we begin by optimizing the feature map. Optimization of the feature map of the second frame requires the lane mask $\bm{M}$ from the first frame. After obtaining the initial features of the second frame through the encoder, we first apply two mapping functions to the second frame feature $\bm{f}_2\in \mathbb{R}^{H\times W\times C}$ to obtain $\bm{k}_2\in \mathbb{R}^{HW\times C}$ and $\bm{v}_2\in \mathbb{R}^{HW\times C}$. The lane mask from the first frame is convolved to generate an ID embedding $E\in \mathbb{R}^{H\times W\times C}$, which is added to $\bm{f}_2$ to obtain $\bm{v}_2$ after the mapping function. The expression is as follows:
\begin{equation}
    \begin{aligned}
    &\bm{E} = ID(\bm{M}), \\
    &\bm{k}_2 = \phi(\bm{f}_2), \\
    &\bm{v}_2 = \varphi(\bm{f}_2 + \bm{E}). \\
    \end{aligned}
\end{equation}
The obtained $\bm{k}_2$ and $\bm{v}_2$ are input into the temporal context aggregation network to produce the optimized features for the second frame.

Starting from the third frame, feature refinement no longer requires the use of a mask. Instead, each frame utilizes the refined features from the preceding frame to optimize the current frame. The refined features of the previous frame are processed through two mapping functions to obtain $\bm{k}_{t-1}$ and $\bm{v}_{t-1}$, which serve as input to the temporal context aggregation network. This approach provides adjacent information.

\begin{figure*}[t]
    \centering
    \includegraphics[width=2\columnwidth]{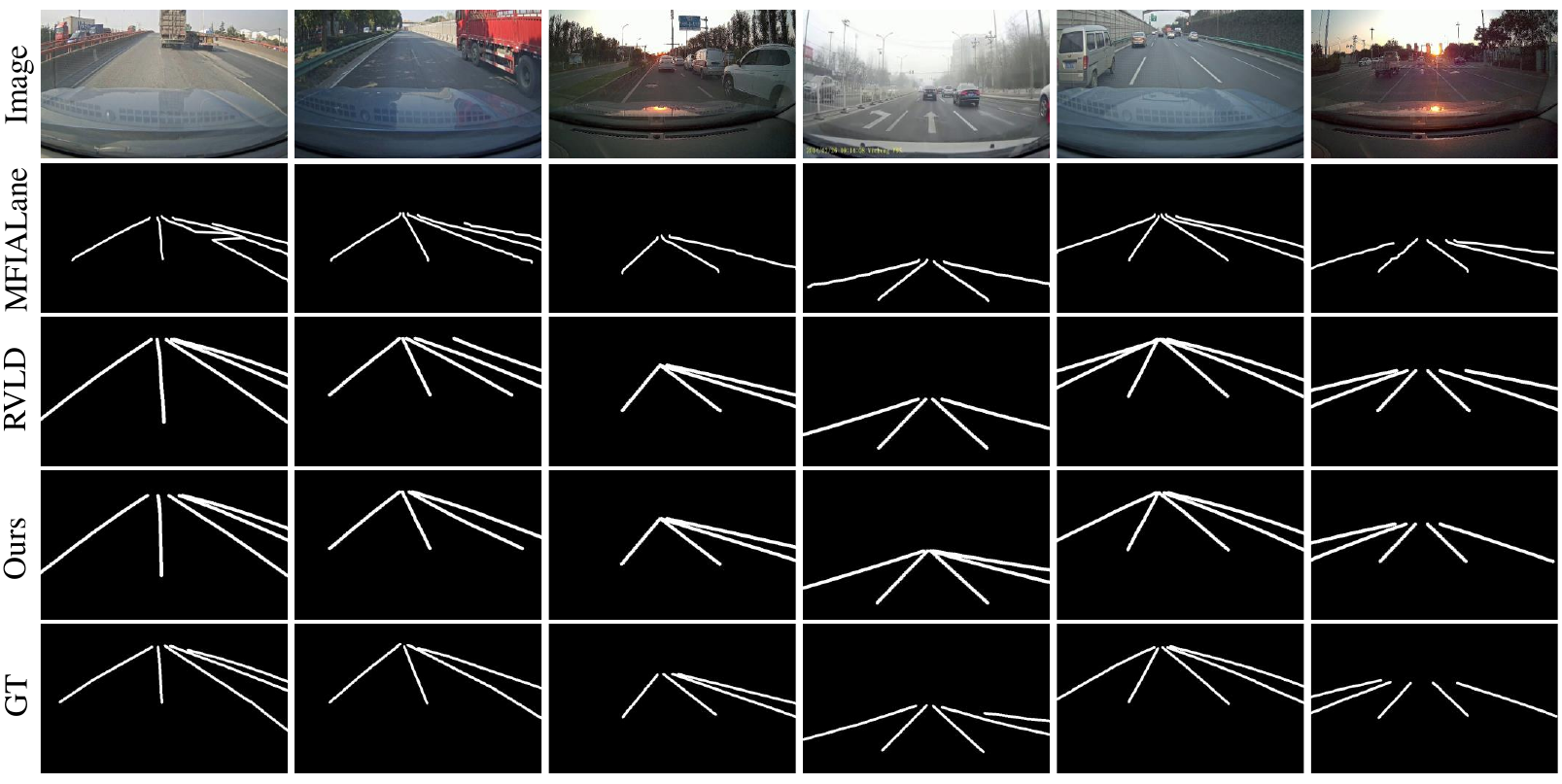}
    \caption{Visualization comparison with state-of-the-art methods on the VIL-100 dataset. From top to bottom, each row represents: the original image, predictions by MFIALane~\cite{qiu2022mfialane}, predictions by RVLD~\cite{jin2023recursive}, predictions by our method, and the ground truth.
    }
    \label{fig:vis-vil}
\end{figure*}

\subsection{Training Objective}

We use loss function $\mathcal{L}$ to constrain the optimized probability map and the final contour of the lane line. For the probability map, we employ focal loss~\cite{lin2017focal} as a constraint. For the parameter map, we utilize the LIoU loss~\cite{zheng2022clrnet} to calculate the constraint. The calculation method is as follows:
\begin{equation}
\mathcal{L} = \mathcal{L}_C(P, \hat{P}) + \mathcal{L}_L(C, \hat{C}),
\end{equation}
where $\hat{P}$ is the ground truth of probability map $P$, and $\hat{C}$ is ground truth of parameter map $C$.

\section{Experiments}

In this section, we first introduce the two two prevalent benchmark datasets used for video lane detection, along with the evaluation metrics. Subsequently, we detail the experimental setup of our study. Following this, we present both quantitative and qualitative results of our LaneTCA, comparing it with other state-of-the-art methods. Finally, we demonstrate the effectiveness of each module and setting of our method through various ablation studies.

\subsection{Datasets}

\smallskip
\textbf{VIL-100.}
The VIL-100 dataset~\cite{zhang2021vil} is a pioneering resource for video lane detection, comprising 100 videos, each with 100 frames. It is divided into 80 videos for training purposes and 20 for testing. This dataset encompasses ten typical driving scenarios, including normal conditions, crowded roads, curved paths, damaged roads, shadows, road markings, dazzling light, haze, nighttime, and crossroads. 

\smallskip
\textbf{OpenLane-V.}
The OpenLane-V dataset~\cite{jin2023recursive} is a substantial resource tailored for video lane detection, derived by refining and re-annotating the original OpenLane dataset~\cite{chen2022persformer}. OpenLane-V~\cite{jin2023recursive} comprises approximately 90,000 images extracted from 590 videos, with a training set of 70,000 images from 450 videos and a test set of 20,000 images from 140 videos. Unlike its predecessor OpenLane~\cite{chen2022persformer}, OpenLane-V~\cite{jin2023recursive} includes annotations for up to four lanes per image as in the CULane dataset, emphasizing the key lanes essential for driving, such as ego and alternative lanes. This enhanced dataset addresses the limitations of OpenLane~\cite{chen2022persformer} by semi-automatically filling in missing lane parts based on matrix completion~\cite{candes2010power}, ensuring temporal consistency and comprehensive lane coverage.

\begin{table}[!t]
    \caption{Results on VIL-100 benchmark dataset~\cite{zhang2021vil}. The best results are highlighted in bold. The second-best results are underlined.}
    \setlength{\tabcolsep}{1mm} 
    \centering
    \label{tab:tableTab}
    \begin{tabular}{clccc} 
        \toprule
         & Method & $mIoU$ $\uparrow$ & $F1^{0.5}$ $\uparrow$ &  $F1^{0.8}$ $\uparrow$ \\
        \midrule
        \multirow{6}*{Image-based} & LaneNet~\cite{neven2018towards}  & 0.633 & 0.721 & 0.222 \\
         & ENet-SAD~\cite{hou2019learning} & 0.616 & 0.755 & 0.205 \\
         & LSTR~\cite{liu2021end} & 0.573 & 0.703 & 0.131 \\
         & RESA~\cite{zheng2021resa} & 0.702 & 0.874 & 0.345 \\
         & LaneATT~\cite{tabelini2021keep} & 0.664 & 0.823 & - \\
         & MFIALane~\cite{qiu2022mfialane} & - & 0.905 & 0.565 \\
         \midrule
        \multirow{5}*{Video-based} & MMA-Net~\cite{zhang2021vil}  & 0.705 & 0.839 & 0.458 \\  
         & LaneATT-T~\cite{tabelini2022lane}  & 0.692 & 0.846 & - \\
         & TGC-Net~\cite{wang2022video} & 0.738 & 0.892 & 0.469 \\
         & RVLD~\cite{jin2023recursive}  & \underline{0.787} & \underline{0.924} & \underline{0.582} \\ 
         & \textbf{Ours}  & \textbf{0.796} & \textbf{0.933} & \textbf{0.621} \\ 
        \bottomrule
        \end{tabular}
     \vspace{-0.01in}
    \label{tab:vil}
\end{table}

\subsection{Evaluation Metrics}

We use F1-score and mIoU to evaluate our method on both VIL-100~\cite{zhang2021vil} and OpenLane-V dataset~\cite{jin2023recursive}. 
Each lane line is considered to have a width of 30 pixels.
When the IoU ratio between the predicted lane marking and the ground truth exceeds the threshold $\tau$, the prediction is considered correct.
True positive (TP), false positive (FP), and false negative (FN) are determined with the IoU threshold of 0.5. 
The formulation is given as follows:
\begin{equation}
    \begin{aligned}
        &F1^{\tau}=\dfrac{2 \times Precision \times Recall}{Precision + Recall},\\
        &Precision=\dfrac{TP}{TP + FP},\\
        &Recall=\dfrac{TP}{TP + FN}.\\
    \end{aligned}
\end{equation}

Additionally, mIoU is determined by averaging the IoU scores of the lanes that have been correctly detected.

\subsection{Implementation Details}

For the encoder, ResNet18~\cite{he2016deep} is appled. For the temporal context aggregation module, we employ MobileNetV2~\cite{sandler2018mobilenetv2} as the backbone.
We set the training epochs as 100 and employ the AdamW optimizer~\cite{loshchilov2017decoupled} for optimization. The initial training rate is $1\times10^{-4}$ for VIL-100 dataset~\cite{zhang2021vil}, and $2\times10^{-4}$ for OpenLane-V dataset~\cite{jin2023recursive} with the weight decay value of $1\times10^{-4}$ for both datasets. 
All the experiments are carried out on the Linux platform using PyTorch with a single NVIDIA GTX 3090Ti GPU.

\begin{figure}[t]
    \centering
    \includegraphics[width=1\columnwidth]{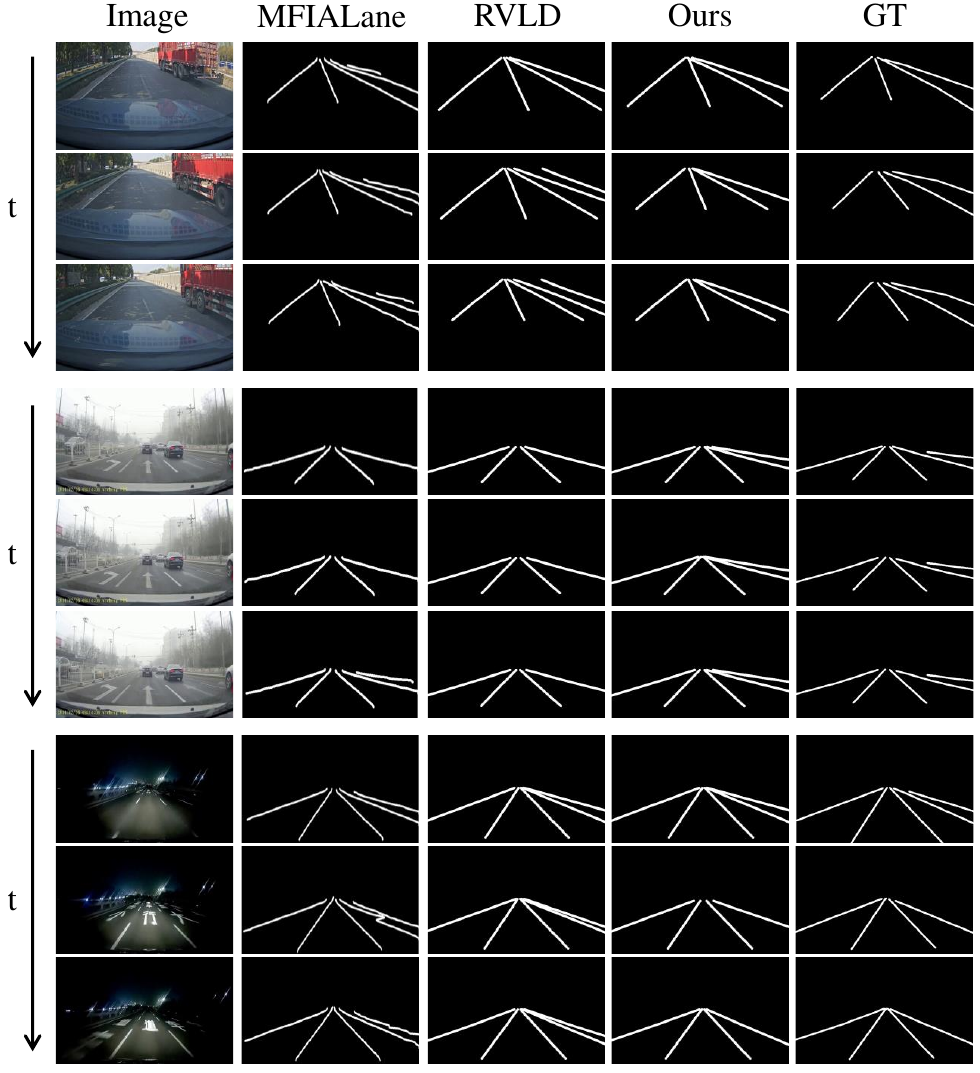}
    \caption{Temporal prediction visualization of our method on the VIL-100 dataset~\cite{zhang2021vil}. From left to right, each column shows: the original image, predictions by MFIALane~\cite{qiu2022mfialane}, predictions by RVLD~\cite{jin2023recursive}, predictions by our method, and ground truth.
    }
    \label{fig:vis-t}
\end{figure}

\subsection{Comparative Assessment}

In the following, We present both quantitative and qualitative results on the VIL-100~\cite{zhang2021vil} and OpenLane-V datasets~\cite{jin2023recursive}.

\smallskip
\textbf{VIL-100.}
We conducted a quantitative comparison with various methods on the VIL-100 dataset~\cite{zhang2021vil}, including LaneNet~\cite{neven2018towards}, ENet-SAD~\cite{hou2019learning}, LSTR~\cite{liu2021end}, RESA~\cite{zheng2021resa}, LaneATT~\cite{tabelini2021keep}, MFIALane~\cite{qiu2022mfialane} as image-based methods, and MMA-Net~\cite{zhang2021vil}, LaneATT-T~\cite{tabelini2022lane}, TGC-Net~\cite{wang2022video}, RVLD~\cite{jin2023recursive} as video-based methods. As presented in Table~\ref{tab:vil}, the results show that our method surpasses all image-based and video-based methods across all metrics.

Compared to image-based methods, we outperform the state-of-the-art MFIALane~\cite{qiu2022mfialane}, demonstrating our effective utilization of temporal information and successful transition from image tasks to video tasks.

For the image-based methods, despite the previous state-of-the-art RVLD~\cite{jin2023recursive} achieving very high performance, our method still surpasses it with improvements of 0.009 in $mIoU$, 0.009 in $F1^{0.5}$, and 0.039 in $F1^{0.8}$. This indicates the superiority of our approach. RVLD~\cite{jin2023recursive} only references the previous frame, whereas we make more extensive use of longer temporal historical information. Even compared to video-based methods that utilize multiple frames, we achieve significant improvements. This is because we combine information from different time spans and are not limited to a fixed number of historical frames, but rather accumulate information from the beginning of the video.

Furthermore, we provide a comparison of our method's visualizations with other approaches in Figure~\ref{fig:vis-vil}. It is shown that our method achieves more accurate results in challenging scenarios such as with occlusion and glare. The temporal visualization is shown in Figure~\ref{fig:vis-t}. It demonstrates that our method exhibits greater temporal continuity and is less prone to sudden prediction errors.

\begin{figure*}[t]
    \centering
    \includegraphics[width=2\columnwidth]{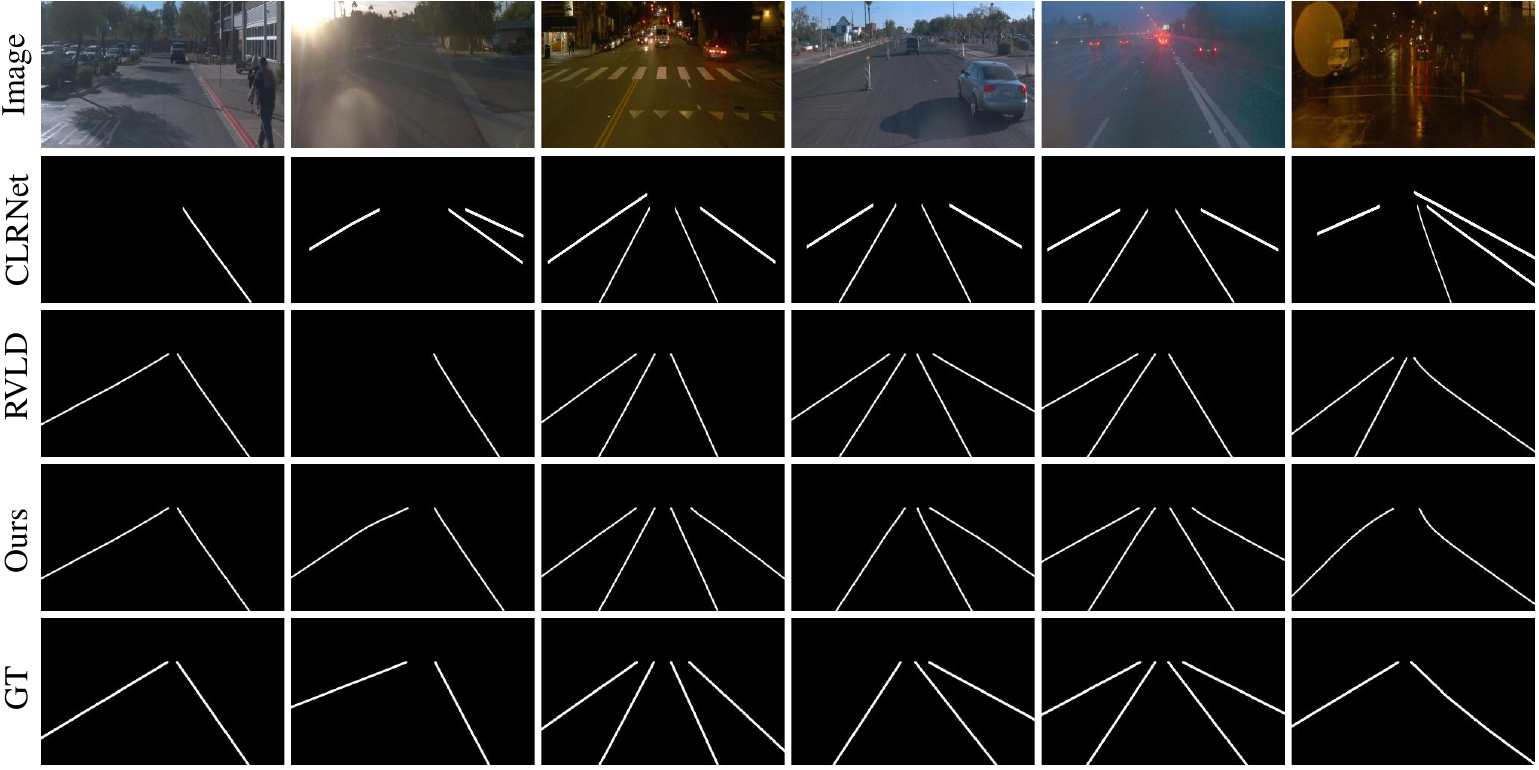}
    \caption{Visualization comparison with state-of-the-art methods on the OpenLane-V dataset~\cite{jin2023recursive}. From top to bottom, each row represents: the original image, predictions by CLRNet~\cite{zheng2022clrnet}, predictions by RVLD~\cite{jin2023recursive}, predictions by our method, and the ground truth.
    }
    \label{fig:vis-ol}
\end{figure*}

\textbf{OpenLane-V.}
We conducted a quantitative comparison of our method with other approaches on the OpenLane-V dataset~\cite{jin2023recursive}, including MFIALane~\cite{qiu2022mfialane}, CondLaneNet~\cite{liu2021condlanenet}, GANet~\cite{wang2022keypoint} and CLRNet~\cite{zheng2022clrnet} as image-based methods, and ConvLSTM~\cite{zou2019robust}, ConvGRUs~\cite{zhang2021lane}, MMA-Net~\cite{zhang2021vil} and RVLD~\cite{jin2023recursive} as video-based methods. The results, presented in Table~\ref{tab:ol}, indicate that our method achieved optimal performance across most metrics. Specifically, compared to the previous state-of-the-art method CLRNet~\cite{zheng2022clrnet}, which achieved the best mIoU, our method significantly surpasses it with an advantage of 0.039 in mIoU, and also exceeds it by 0.033 and 0.02 in $F1^{0.5}$ and $F1^{0.8}$, respectively.

For RVLD~\cite{jin2023recursive}, which previously achieved the best performance in $F1^{0.5}$ and $F1^{0.8}$, our method surpasses it by 0.047 in mIoU and also by 0.008 in $F1^{0.8}$. It is noteworthy that among the methods compared on the OpenLane-V dataset~\cite{jin2023recursive}, most image-based methods demonstrate superior performance, surpassing the accuracy of many video-based methods. This suggests that the effectiveness of temporal information utilization is not evident on OpenLane-V~\cite{jin2023recursive}. Nonetheless, our method still outperforms all image-based methods, demonstrating our method's effective utilization of temporal information.
The visualization is shown in Figure~\ref{fig:vis-ol}.

\begin{table}[!t]
    \caption{Results on OpenLane-V benchmark dataset~\cite{jin2023recursive}. The best results are highlighted in bold. The second-best results are underlined.}
    \setlength{\tabcolsep}{1.3mm} 
    \centering
    \label{tab:tableTab}
    \begin{tabular}{clccc} 
        \toprule
         & Method & $mIoU$ $\uparrow$ & $F1^{0.5}$ $\uparrow$ &  $F1^{0.8}$ $\uparrow$ \\
        \midrule
        \multirow{4}*{Image-based} & MFIALane~\cite{qiu2022mfialane} & 0.697 & 0.723 & 0.475 \\
         & CondLaneNet~\cite{liu2021condlanenet} & 0.698 & 0.780 & 0.450 \\
         & GANet~\cite{wang2022keypoint} & 0.716 & 0.801 & 0.530 \\
         & CLRNet~\cite{zheng2022clrnet} & \underline{0.735} & 0.789 & 0.554 \\
         \midrule
        \multirow{5}*{Video-based} & ConvLSTM~\cite{zou2019robust}  & 0.529 & 0.641 & 0.353 \\  
         & ConvGRUs~\cite{zhang2021lane}  & 0.540 & 0.641 & 0.355 \\
         & MMA-Net~\cite{zhang2021vil}  & 0.574 & 0.573 & 0.328 \\
        & RVLD~\cite{jin2023recursive}  & 0.727 & \textbf{0.825} & \underline{0.566} \\ 
         & \textbf{Ours}  & \textbf{0.774} & \underline{0.822} & \textbf{0.574} \\ 
        \bottomrule
        \end{tabular}
     \vspace{-0.01in}
    \label{tab:ol}
\end{table}

\subsection{Ablation Studies}

In this section, we perform extensive experiments to validate the
core components and settings of our method, including adjacent attention, accumulative attention, temporal context aggregation network (TCA), and mask cue setting. All the ablations are conducted on the benchmark VIL-100 dataset~\cite{zhang2021vil}.

\smallskip
\textbf{Impact of the adjacent attention.}
Adjacent attention is employed to provide highly correlated information from neighboring frames. We conducted experiments to verify the effectiveness of the adjacent attention. We performed ablation studies by removing the accumulative attention block. The results in Table~\ref{tab:acl} demonstrated that the adjacent attention improves detection accuracy. This indicates that the highly relevant information provided by adjacent frames contributes positively to the prediction of the current frame.

\smallskip
\textbf{Impact of the accumulative attention.}
Accumulative attention is utilized to provide long-term information that has been continuously learned from the beginning of the video. We conducted experiments to verify the effectiveness of the accumulative attention by removing the adjacent attention block. The results demonstrated that the accumulative attention improves detection accuracy. This indicates that the long-term information accumulated from the beginning contributes positively to the prediction of the current frame.

We validated the effectiveness of accumulative attention from another perspective. We aimed to explore whether this cumulative learning approach, which breaks the limitation of a fixed number of frames, is indeed superior to using a fixed number of frames for each individual frame. To this end, we conducted experiments by setting different time spans for accumulative attention.
Initially, the default setting was to initialize the learnable query of accumulative attention from the beginning of the video and continuously learn road information. By adjusting the accumulative length, we set the learnable query to be initialized at different fixed frame intervals. The results, shown in Table~\ref{tab:acl}, indicate that as the fixed frame number increases, the detection accuracy gradually improves and approaches the result of continuously accumulating road information from the beginning. This demonstrates that accumulative attention, which continuously learns from the start of the video, is effective.

\begin{table}[!t]
    \caption{Ablation study on the adjacent attention and accumulation attention.}
    \setlength{\tabcolsep}{1mm} 
    \centering
    \label{tab:tableTab}
    \begin{tabular}{cccc} 
        \toprule
        {Adjacent Attention} & {Accumulative Attention} & $mIoU$ $\uparrow$ & $F1^{0.5}$ $\uparrow$ \\
        \midrule
          &  & 0.768 & 0.923 \\
        \checkmark &  & 0.773 & 0.928 \\
           & \checkmark & 0.786 & 0.930 \\
        \checkmark  & \checkmark  & \textbf{0.796} & \textbf{0.933} \\
        \bottomrule
        \end{tabular}
     \vspace{-0.01in}
    \label{tab:att}
\end{table}

\begin{table}[!t]
    \caption{Ablation study on the accumulative length. The settings \underline{underlined} are used in final model.}
    \setlength{\tabcolsep}{2mm} 
    \centering
    \label{tab:tableTab}
    \begin{tabular}{ccc}  
        \toprule
        {Accumulative Length} & $mIoU$ $\uparrow$ & $F1^{0.5}$ $\uparrow$ \\
        \midrule
        4 & 0.784 & 0.931 \\
        8 & 0.787 & 0.932 \\
        16 & 0.790 & 0.933 \\
        \underline{all frames} & \textbf{0.796} & \textbf{0.933} \\
        \bottomrule
        \end{tabular}
     \vspace{-0.01in}
    \label{tab:acl}
\end{table}

\smallskip
\textbf{Impact of TCA.}
The structure of TCA integrates the information provided by adjacent attention and accumulative attention with the information obtained from the current frame. The optimized features derived from this integration enable more accurate prediction of lane lines at the present moment. The results in Table~\ref{tab:acl} demonstrate that the use of adjacent attention and accumulative attention simultaneously produces better outcomes compared to using either one alone. This indicates that these two mechanisms do not inhibit each other, but rather provide complementary short-term and long-term information, effectively enhancing the current feature set.

\begin{figure}[t]
    \centering
    \includegraphics[width=1\columnwidth]{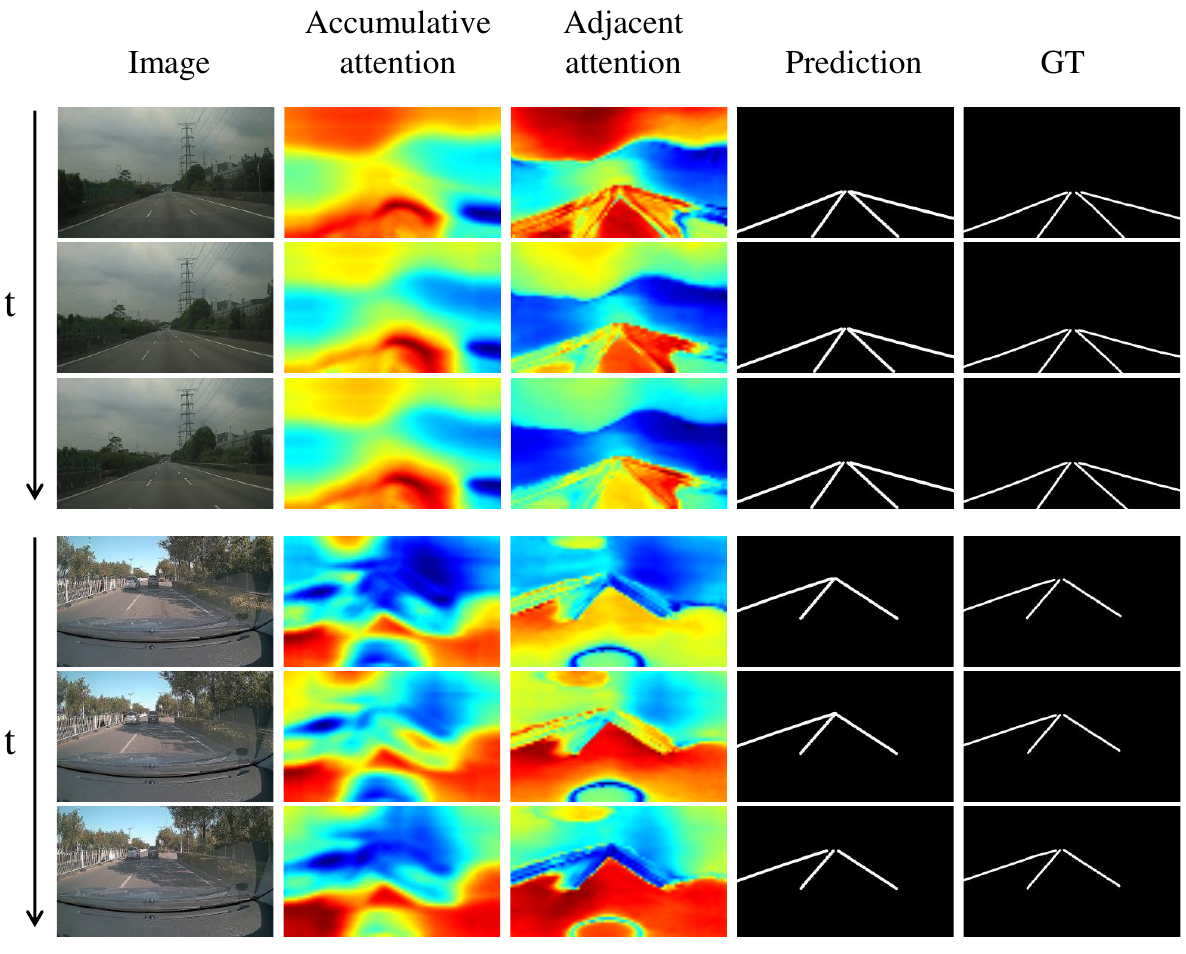}
    \caption{Attention visualization of our method on the VIL-100 dataset~\cite{zhang2021vil}. From left to right, each column shows: the original image, accumulative attention, adjacent attention, predicted result, and ground truth.
    }
    \label{fig:vis-att}
\end{figure}

We attempt to explore the interpretability of our LaneTCA in improving the performance of video lane detection. Figure~\ref{fig:vis-att} presents the temporal visualization results of accumulative attention and adjacent attention. The results shown here are derived from the outputs of both the accumulative attention and adjacent attention modules, i.e., before the two branches are combined. We observe that these two types of attention exhibit complementary phenomena. Specifically, accumulative attention demonstrates a focus on broader areas, serving as an indicator of motion trends after long-term information learning. In contrast, adjacent attention shows a higher level of distinction for detailed information, emphasizing judgments based on very short time spans. The addition of these relatively coarse and fine attention results can integrate information at different scales at the feature level, leading to the more accurate outcomes.

From a temporal perspective, accumulative attention exhibits minor changes between adjacent frames, aligning with real-world scenarios where the time difference between adjacent frames is minimal and the road scene changes very little. However, adjacent attention may exhibit abrupt changes between adjacent frames. In such cases, combining it with accumulative attention provides a smoothing effect. Therefore, it is evident that both attention branches hold significant and specific importance in the detection process.

\smallskip
\textbf{Impact of the mask cue.}
The mask cue is used to provide supplementary information during the initial phase of a video. To verify the effectiveness of the mask cue, we tested several configurations: not using a mask cue, using a mask cue throughout, and using the mask cue of the first frame only when predicting the second frame. Experimental results shown in Table~\ref{tab:msk} indicate that detection without a mask cue outperforms using a mask cue throughout, while using the mask cue only for the second frame produces the best results among the three configurations. Our analysis suggests that in long video detection tasks, the lane mask, being a slender structure, loses its effectiveness as a cue over time. Additionally, in scenarios involving vehicle lane changes or sudden reductions in the number of lane markings, consistently using the mask cue can be misleading. However, during the second frame detection at the start of the video, the accumulative attention has accumulated very limited information. Thus, the mask cue at this stage can compensate for the lack of long-term information during the initial phase.

\begin{table}[!t]
    \caption{Ablation study on the setting of mask cue. The settings \underline{underlined} are used in final model.}
    \setlength{\tabcolsep}{1mm} 
    \centering
    \begin{tabular}{ccc} 
        \toprule
        {Setting} & $mIoU$ $\uparrow$ & $F1^{0.5}$ $\uparrow$ \\
        \midrule
        w/o & 0.787 & 0.929 \\
        for all frame & 0.784 & 0.921 \\
        \underline{for second frame} & \textbf{0.796} & \textbf{0.933} \\
        \bottomrule
        \end{tabular}
     \vspace{-0.01in}
    \label{tab:msk}
\end{table}

\section{Conclusion}

In this work, we propose a novel video lane detection network that integrates features from both adjacent and long-term information to achieve improved current frame detection results. We design the adjacent attention block to capture short-term features by incorporating information from the previous frame. Additionally, we design the accumulative attention block to continuously integrate information from each frame, thus obtaining long-term features without frame number limitations. Experimental results demonstrate that the combination of these two modules significantly enhances video lane detection performance. Our method achieves competitive results compared to the existing methods on the VIL-100 and OpenLane-V datasets.

{
\bibliographystyle{IEEEtran}
\bibliography{IEEEabrv,egbib}
}

\end{document}